\documentclass{article}

\usepackage[preprint]{neurips_2025}

\usepackage[utf8]{inputenc} % allow utf-8 input
\usepackage[T1]{fontenc}    % use 8-bit T1 fonts
\usepackage{hyperref}       % hyperlinks
\usepackage{url}            % simple URL typesetting
\usepackage{booktabs}       % professional-quality tables
\usepackage{amsfonts}       % blackboard math symbols
\usepackage{nicefrac}       % compact symbols for 1/2, etc.
\usepackage{microtype}      % microtypography
\usepackage{xcolor}         % colors
\usepackage{graphicx} % 加载图形包

\title{Formatting Instructions For NeurIPS 2025}
\title{Learning Compositional Transferability of Time Series for Source-Free Domain Adaptation}

\author{%
  Hankang Sun \\
  Fudan University\\
  \texttt{23210240277@m.fudan.edu.cn} \\
  \And
  Guiming Li \\
  Fudan University\\
  \texttt{23210240210@m.fudan.edu.cn} \\
  \AND
  Su Yang\thanks{Corresponding author} \\
  Fudan University\\
  \texttt{suyang@fudan.edu.cn} \\
  \And
  Baoqi Li \\
  Chinese Academy of Sciences, Institute of Acoustics \\
  \texttt{libaoqi@mail.ioa.ac.cn} \\
}

\begin{document}

\maketitle

\begin{abstract}
   Domain adaptation is challenging for time series classification due to the highly dynamic nature. This study tackles the most difficult subtask when both target labels and source data are inaccessible, namely, source-free domain adaptation. To reuse the classification backbone pre-trained on source data, time series reconstruction is a sound solution that aligns target and source time series by minimizing the reconstruction errors of both. However, simply fine-tuning the source pre-trained reconstructor on target data may lose the learnt priori from source data, which plays an important role in the overall adaptation as observed by us, and furthermore, it struggles to accommodate domain varying temporal patterns in a single encoder-decoder. Therefore, this paper tries to disentangle the composition of domain transferability by using a compositional architecture for time series reconstruction, instead of fine-tuning a single encoder-decoder. Here, the preceding component is a U-net frozen since pre-trained, the output of which during adaptation is the initial reconstruction of a given target time series, acting as a coarse step to prompt the subsequent finer adaptation. The following pipeline for finer adaptation includes two parallel branches: The source replay branch using a residual link to preserve the output of U-net, and the offset compensation branch that applies an additional autoencoder (AE) to further warp U-net’s output. By deploying a learnable factor on either branch to scale their composition in the final output of reconstruction, the data transferability is disentangled and the learnt reconstructive capability from source data is retained. During inference, aside from the batch-level optimization in the training, we search at test time stability-aware rescaling of source replay branch to tolerate instance-wise variation. The experimental results\footnote{Our experiment code is available at \url{https://anonymous.4open.science/r/CT-SFDA-Code/}} show that such compositional architecture of time series reconstruction leads to SOTA performance with 2.68\%, 0.23\%, and 2.36\% improvement on 3 widely used benchmarks. After applying the instance-wise rescaling, the improvement goes further up to 3.7\%, 0.78\%, and 2.6\%.
\end{abstract}

% 引入
\section{Introduction}
Time series classification plays an important role in a variety of applications \citet{wang2023wavelet,gorbett2023sparse,mingyue2023formertime}, such as human activity recognition using the data from wearable sensors \cite{xu2023channel,hu2023swl,kang2024sf,ye2024deep} , mechanical fault diagnosis \cite{tian2024universal,luo2024fft,qian2024adaptive}, and EEG classification \cite{zhao2020deep,pradeepkumar2024towards,zhang2024multi}. So far, a major challenge for applying time series classification in practice is context-related domain shift. For example, a pre-trained model for mechanical fault diagnosis may degrade significantly when working under varied rolling speed or torque in contrast to the pre-training condition. This leads to the raising interest in domain adaptation, which aims to reuse pre-trained classification models by aligning target and source domains or learning universal representations. To date, domain adaptation for time series is yet a challenging issue due to the highly dynamic nature of time series. According to the difficulty from low to high, the domain adaptation tasks for time series classification can be sorted into 3 settings: Few-shot learning, unsupervised domain adaptation (UDA), and source-free domain adaptation (SFDA). This study is focused on SFDA, which is the most difficult scenario undergoing less investigations since both source data and target labels are inaccessible during adaptation. 

\begin{figure*}[t]
\centering
\includegraphics[width=0.6\textwidth]{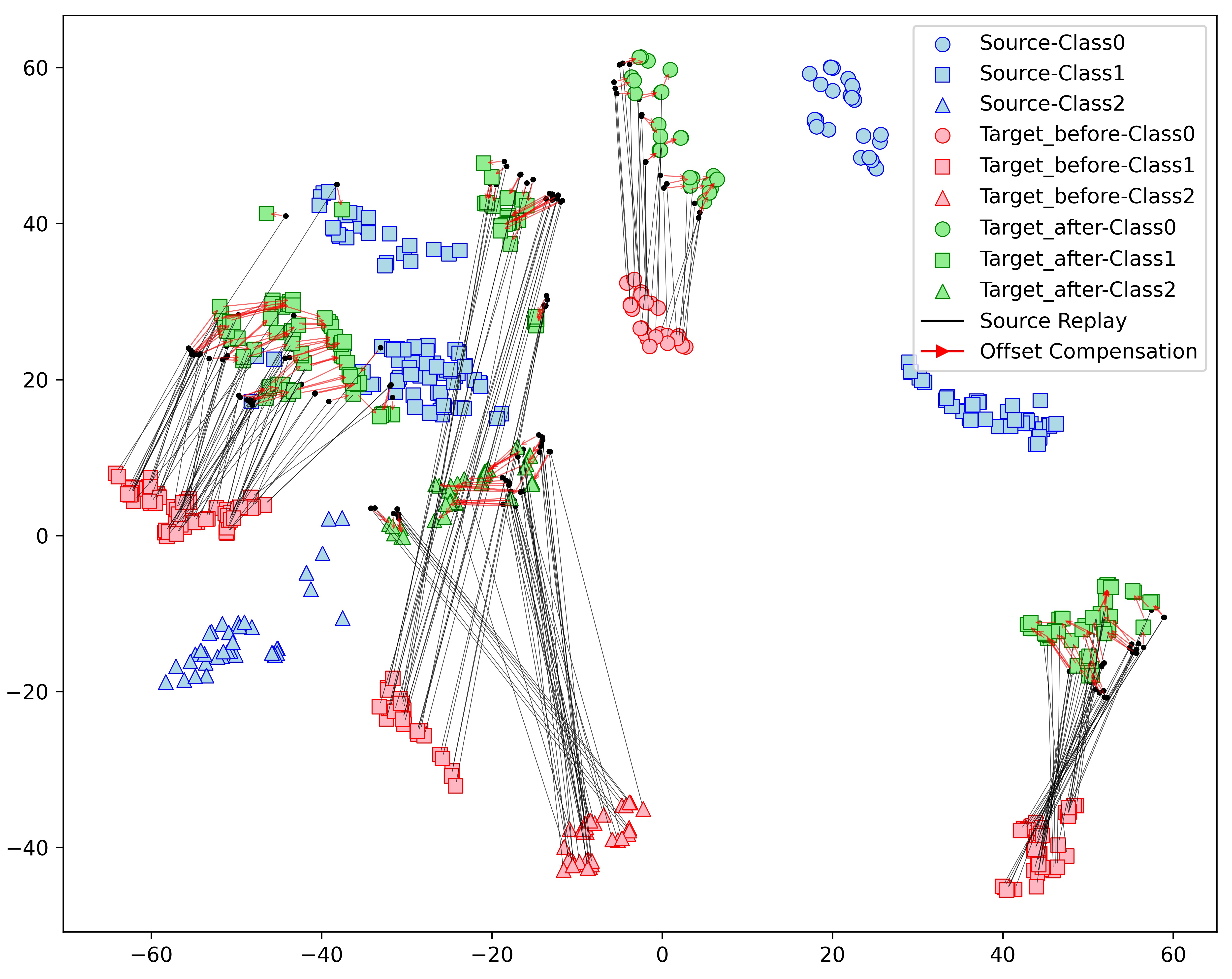} % Reduce the figure size so that it is slightly narrower than the column.
\caption{Composition of Domain Transferability (The latent features are illustrated in different shapes to denote their belongings to 3 classes, where circle, square, and triangle represent the data instances of class 0, class 1, and class 2 in the MFD data, respectively. Further, blue, red, or green color is assigned to each data instance to denote the source data, the target data before adaptation, and the target data after adaptation, where the black lines indicate the course of migrating the target data from their original positions to the transition points following source replay, marked using black dots, and the subsequent migrating from such transition positions to the destinations after applying additional offset compensation, denoted using red lines). The domain discrepancy is progressively shortened for each class following source replay and offset compensation in order.}
\label{fig1}
\end{figure*}

In the literature, time series reconstruction is a sound solution for domain alignment, which uses reconstruction error as a measure of alignment. As a result, the reconstructed time series can be applied as input to the pre-trained classification model. Since the adaptation operates on time series rather than latent space, it does not require knowing the internal structure of classification backbone, making it universally applicable. However, the existing work \cite{he2023domain} utilizes a single encoder-decoder for time series reconstruction and approaches domain alignment by using target data to fine-tune the reconstruction module pre-trained on source data. Such a strategy will lose the priori learnt from source data, which is informative and can act as anchors to direct domain alignment. Besides, tuning a single encoder-decoder to accommodate domain varying temporal patterns is sometimes out of its capability on account of the highly dynamic nature of the data, resulting in compromised solution.

\begin{table}[h]
\centering
\caption{Comparison of two methodologies. }
\label{tab:tta}
\begin{tabular}{l|ccc}
\toprule
MF1 score(\%) & {MFD} & {SSC} & {UCIHAR} \\
\midrule
Fine-tune only & 91.37 & 63.04 & 86.47 \\
Two-phase transfer   & \textbf{95.13} & \textbf{64.28} & \textbf{91.93} \\
\bottomrule
\end{tabular}
\end{table}

In view of the limit of single encoder-decoder, we develop a compositional architecture for time series reconstruction towards domain alignment. The leading component is a U-net frozen since pre-trained on source data. The following pipeline includes two parallel branches: The source replay branch that preserves the priori learnt from pre-training by deploying a residual link, and the offset compensation branch that applies an additional autoencoder(AE) to warp the time series tuned from U-net for further reducing the discrepancy across domains. The overall reconstruction combines the outputs of the two branches. As the visualized example illustrated  in Figure 1, target domain transferring can thus be decomposed into two compositional parts: (1) The solid black dots denoting the transition positions in the course of the overall target transferring, obtained by applying the target time series to the frozen pre-trained U-net to yield source priori-informed intermediate transferring; (2) The subsequent migrating from the black dot-denoted transition positions to the destinations close to source data by using the dual-branch postprocessing. We can see that the black dot-represented intermediately reconstructive time series from U-net correspond with the majority part of the whole journey of data transferring to reduce domain gap, and the subsequent offset compensation on the basis of source replay can further shorten the domain gap at a finer scale. This means that the prior knowledge inherited by source replay, that is, the black dots in Figure 1, plays an important role in the overall domain transferring from target to source. In fact, source replay acts as a platform in correspondence with essentially coarse adaptation, and offset compensation advances towards finer adaptation based on this platform as onset. Comparison between this two-phase transferring and fine-tuning pre-trained reconstruction module on 3 benchmarks is summarized in Table 1. We can see that the aforementioned compositional transferring outperforms solely fine-tuning the pre-trained U-net. We attribute this to using the frozen pre-trained model as a priori but fine-tuning the source model fails to preserve such priori. Since source replay branch plays an important role in composing the overall adaptation as shown previously, we advocate the compositional time series reconstruction and choose U-net as the backbone for source reconstruction in order to take advantage of its strong data fitting power resulting from the large number of learnable parameters, which has been verified in many computer vision scenarios. Here, the time series is reshaped to an image-like form prior to being fed to U-net such that the strong model can be applied directly to tackle time series. The advantage of treating time series as image has been well investigated in \cite{li2024time}, where they use Transformer for a different task. 

Aside from the previously described group-level adaptation, another issue is instance-wise adaptation, which has been less studied in the literature but is of importance. Here, we propose to perform further instance-wise adaptation by rescaling the source replay branch at test time to test its impact on the stability of classification, and ensemble classifier’s outputs weighted by such stability test.

The experimental results show that such compositional architecture of time series reconstruction leads to the state-of-the-art (SOTA) performance with 2.68\%, 0.23\%, and 2.36\% improvement on 3 widely used benchmarks. After applying the instance-wise rescaling, the improvement goes further up to 3.7\%, 0.78\%, and 2.6\%.

The contribution of this study is summarized as follows:
\begin{itemize}
\item A compositional architecture of time series reconstruction is developed to combine replayed source priori and offset compensation for domain adaptation. 
\item A group-to-instance adaptation strategy is applied by combining classifier’s outputs based on stability test on the fly.
\item The experiments show that the proposed solution can achieve SOTA performance on 3 benchmarks with 3.7\%, 0.78\%, and 2.6\% increment in terms of MF1 score.
\end{itemize}

% 相关工作
\section{Related Works}

\subsection{Time Series Domain Adaptation}
Upon the different scenarios in terms of data availability, domain adaptation (DA) has the problem settings as follows: General DA uses few-shot labels to train the adapter or prompt big models for downstream adaptation \cite{wang2024pond,kang2024sf,zhang2022self}. In the unsupervised domain adaptation (UDA) setting \cite{he2023domain}, unlabeled target data along with labeled source data are available. Source-free domain adaptation (SFDA) has the most rigorous setting to allow access to unlabeled target data only, while accessing to source data or labels is prohibited \cite{ragab2023source}. We follow the setting of MAPU \cite{ragab2023source} to assume that source and target domains share the same categories with feature distribution shift only, and the pre-trained modules are accessible, but excluding source data. Although SFDA has undergone dense investigations in computer vision, SFDA for time series classification is far from matured on account of the recently developed benchmark \cite{ragab2023adatime}. 

\subsection{Reconstructive Alignment}
Encoder-decoder is a well demonstrated architecture for domain alignment, which turns the problem into time series reconstruction without accessing to class labels, thus very suitable for UDA and SFDA. Moreover, as the adaptation operates on time series rather than latent space, it does not require knowing the internal structure of classification backbone, making it universally applicable. The early effort stems from computer vision: DRCN \cite{ghifary2016deep} applies a shared encoder for both reconstruction and classification but the multi-task coupling makes model training tougher. RAINCOAT \cite{he2023domain} trains a single encoder-decoder to align source and target time series based on the assumption that normal data examples of both domains are homogeneous or overlap substantially to allow yielding a uniquely  well-fitted model for time series reconstruction. MAPU \cite{ragab2023source} enforces an autoregressive network to reconstruct partially masked time series by fine-tuning the encoder pre-trained on source domain with target data. These methods assume implicitly that compatible dynamics exist between source and target, and can thus be captured by a single reconstructive module for alignment. In practice, however, the highly complex dynamics of time series make the global as well as transient patterns less regular, and not easy to be learnt by a single model. This study proposes a compositional architecture, aiming to fit better into the heterogeneous composition of time series across domains.

\subsection{Universal Representation Learning}
Another sound methodology for domain alignment seeks domain-invariant features via representation learning \cite{liu2024timesurl}. CALDA \cite{wilson2023calda} trains a domain classifier under adversarial loss along with domain-aware contrastive learning to make the encoder domain invariant. AdvSKM \cite{liu2021adversarial} maps cross-domain data into a universal feature space by training shared encoder, along with adversarial training in spectral kernel space to align data examples. MHCCL \cite{meng2023mhccl} applies multi-level contrastive losses to enhance representation learning. SASA \cite{cai2021time} aims to align invariant local structure by summarizing locally associated patterns of a time series into domain-invariant representations, based on learnable attention weights and LSTM encoder. Nevertheless, these approaches struggle to accommodate heterogeneous patterns using an identical encoder, while its complex coupling with classification backbone makes the task tougher.

\subsection{Test Time Adaptation}
Test time adaptation (TTA) has received much attention in computer vision but rarely been studied for time series classification. TENT \cite{wang2020tent} fine-tunes batch normalization (BN) layers at test time using selective data examples outlying in a batch. TTN \cite{lim2023ttn} combines conventional BN and target BN to tackle domain shift caused degrading. Sharpness-aware entropy \cite{niu2023towards} aims to stabilize classification by examining the sensitivity of parameters against perturbation. SoTTA \cite{gong2024sotta} also addresses parameter-wise robustness against noisy samples via entropy-sharpness minimization. However, the existing TTA methods mostly rely on batch-wise update or updating a large number of parameters , making the solutions inefficient. This study proposes an efficient TTA method to augment our reconstructive module by deploying instance-wise sliding test to rescale source replay, for the sake of approaching robust time series classification in an ensemble manner.

\begin{figure*}[t]
\centering
\includegraphics[width=\textwidth]{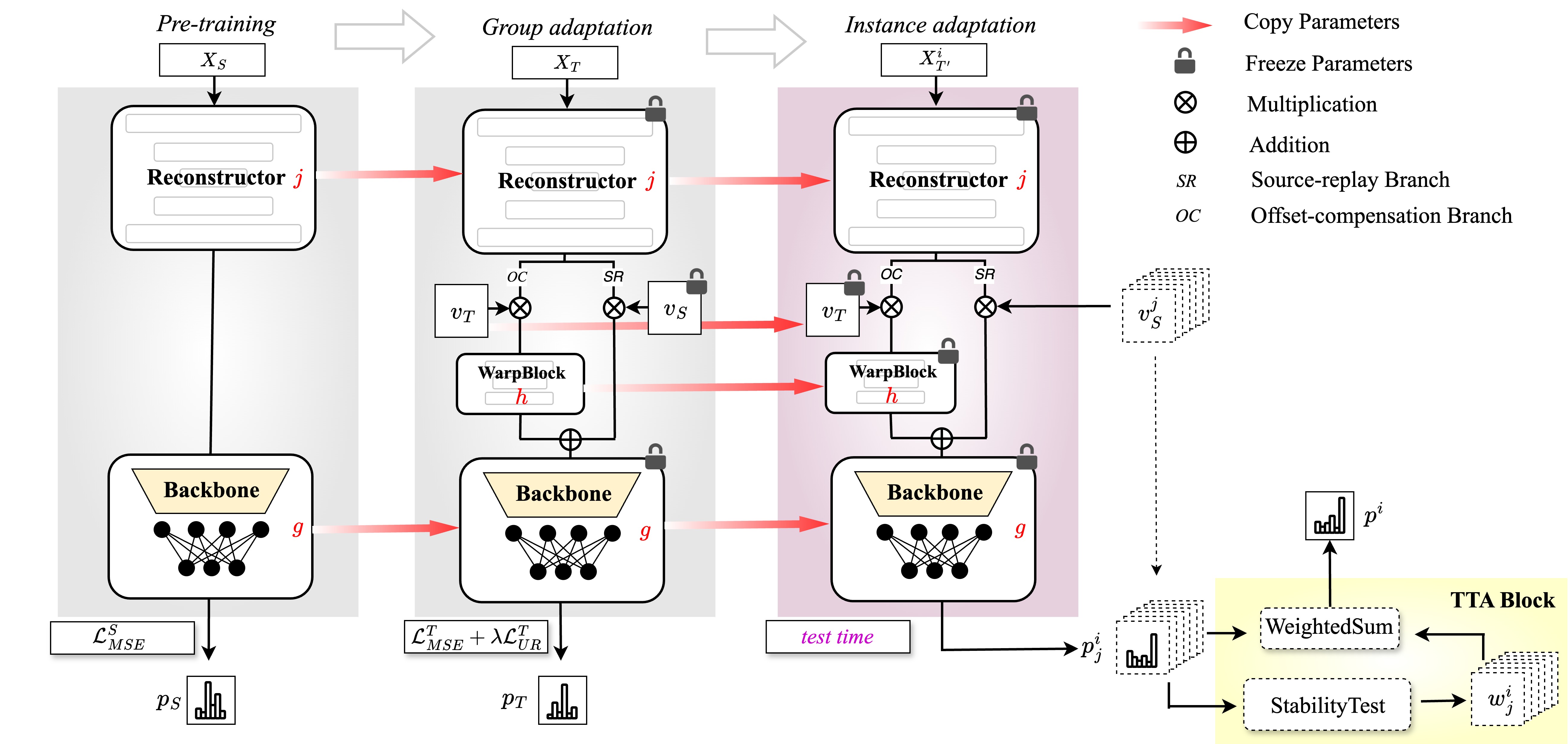} % Reduce the figure size so that it is slightly narrower than the column.
\caption{Network architecture.}
\label{fig2}
\end{figure*}

\section{Compositional Time Series Reconstruction}

In regard to Figure 1 and Table 1, we propose a compositional architecture for time series reconstruction, by replaying source model at first to yield informative priori in the entire course of transferring, and solving the remaining task as offset compensation on the basis of source replay. The detailed implementation is described below.

\subsection{Preliminaries}

${\mathcal D}_{S} = \left\{X_{S}^{i}, y_{S}^{i}\right\}_{i = 1}^{N_{S}}$ denotes a labeled dataset from source domain, where  $X_S^i\in {\mathbb R}^{d\times L}$ represents uni-variate $(d=1)$ or multi-variate $(d>1)$ time series of length $L$, and $y_S^i\in {\mathbb R}^{K}$ the corresponding labels. ${\mathcal D}_{T} = \left\{X_{T}^{i}\right\}_{i = 1}^{N_{T}}$ represents a unlabeled dataset from target domain, which shares the same label space $y=\{1,2,\cdots,K\}$ with ${\mathcal D}_{S}$. We follow the common settings of SFDA to assume that the marginal distributions ${P}_{S}({X}_{S})\ne{P}_{T}({X}_{T})$ due to feature shift, and it is strictly prohibited to access source data in the adaptation phase. During test time, we are given batch-wise unlabeled target instances $\{X_{{T}'}^{i}\}_{i=1}^B{(B\ge1)}$ from the same distribution of ${\mathcal D}_{T}$. Note that $B>1$ is totally for parallel computing, not affecting instance-wise adaptation.

\subsection{Network Architecture}
The components that need to be pre-trained before adaptation include the backbone network for time series classification and the time series reconstructor based on encoder-decoder architecture. Both are trained on source domain data and kept frozen since then. Here, we choose U-net as the reconstructive backbone due to its strong power for data fitting, powering the subsequent source replay branch to provide the platform for successively finer adaptation. As demonstrated, treating time series as images allows direct use of strong computer vision models \cite{li2024time}.

As shown in Figure 2, we design a dual-branch network structure. During adaptation, a target domain time series is firstly processed by the preceding reconstructor learnt from source data, and then passed through two parallel branches, namely, the source replay branch and the offset compensation branch, to obtain synthetic result, which is finally fed to the source pre-trained backbone for classification. The offset compensation branch pipelines sequentially two trainable components: A warp block based on autoencoder and a coupled scaling factor to weight the warping. The source replay branch directly inherits the output from the source pre-trained reconstructor with a fixed scaling factor of 1 at the group-level adaptation phase. 

Instance adaptation is deployed during inference. For an input instance, we adjust the scaling factor of the source replay branch on the fly. By sliding search within the neighborhood around the initial value of 1, we aggregate the classifier’s outputs weighted by a measure of stability against such small perturbation on the scaling factor.

This pipeline allows incremental knowledge learning and test-time adaptation, in a coarse-to-fine, fundamental-to-specific, and group-wise to instance-aware manner.And the paradigm allows a complicate task to be disentangled into easier and lightweight subtasks.

\subsection{Pre-training on Source Domain}
As shown in Figure 2, we pre-train a classification backbone and an accompanying time series reconstructor based on source data, where the reconstructed time series serves as the input to the classification backbone. Here, U-net is applied as the reconstructor, aiming to learn the distribution of the source domain data implicitly, based on its strong data fitting power arising from the large number of learnable parameters. The reconstruction loss to be optimized is as follows:

\begin{equation}
\min_{\theta} {\mathcal L}_{MSE}^{S} = \frac{1}{N_S} \sum_{i=1}^{N_S} \left \|X_{S}^{i} - {j}_\theta(X_{S}^{i})\right \|^2
\end{equation}
where $j_\theta$ represents the reconstructor, and $\theta$ denotes the parameters to be optimized in the sense of mean squared error (MSE), frozen since pre-trained.

As demonstrated in MAPU \cite{ragab2023source}, one-dimensional convolutional neural network (1D-CNN) performs well in time series classification tasks. For the sake of comparison, we use the same network structure as applied in MAPU \cite{ragab2023source} to train the source domain classification backbone, which consists of a 1D-CNN encoder and a classifier. We apply the output from the pre-trained reconstructor as the input to the downstream classification backbone and minimize the cross entropy loss in training. 

\subsection{Group-level Adaptation}

As shown in Figure 2, the target time series is firstly fed to the frozen reconstructor (U-net), and then the output of U-net is further applied to the dual branches to undergo subsequent processing in order to minimize the reconstruction error on the target domain. The resulting sequence is the direct sum of the two branches’ outputs, which is applied to the source pre-trained classification backbone. Note that our method operates on time series rather than latent-space features, so it does not require knowing the internal structure of the backbone network, making it widely applicable to various downstream tasks. 

\subsubsection{Source Replay Branch.} During the pre-training, the U-net based reconstructor has sufficiently learnt the distribution of source data. Therefore, we use a residual link to let the target sequence reconstructed from U-net compose partially the final reconstructive output, namely $j_\theta\left(X_T^i\right)$. This can be regarded as a replay of the source model with learnt knowledge frozen in U-net.

\begin{figure}[h]
  \centering
  \includegraphics[width=0.6\linewidth]{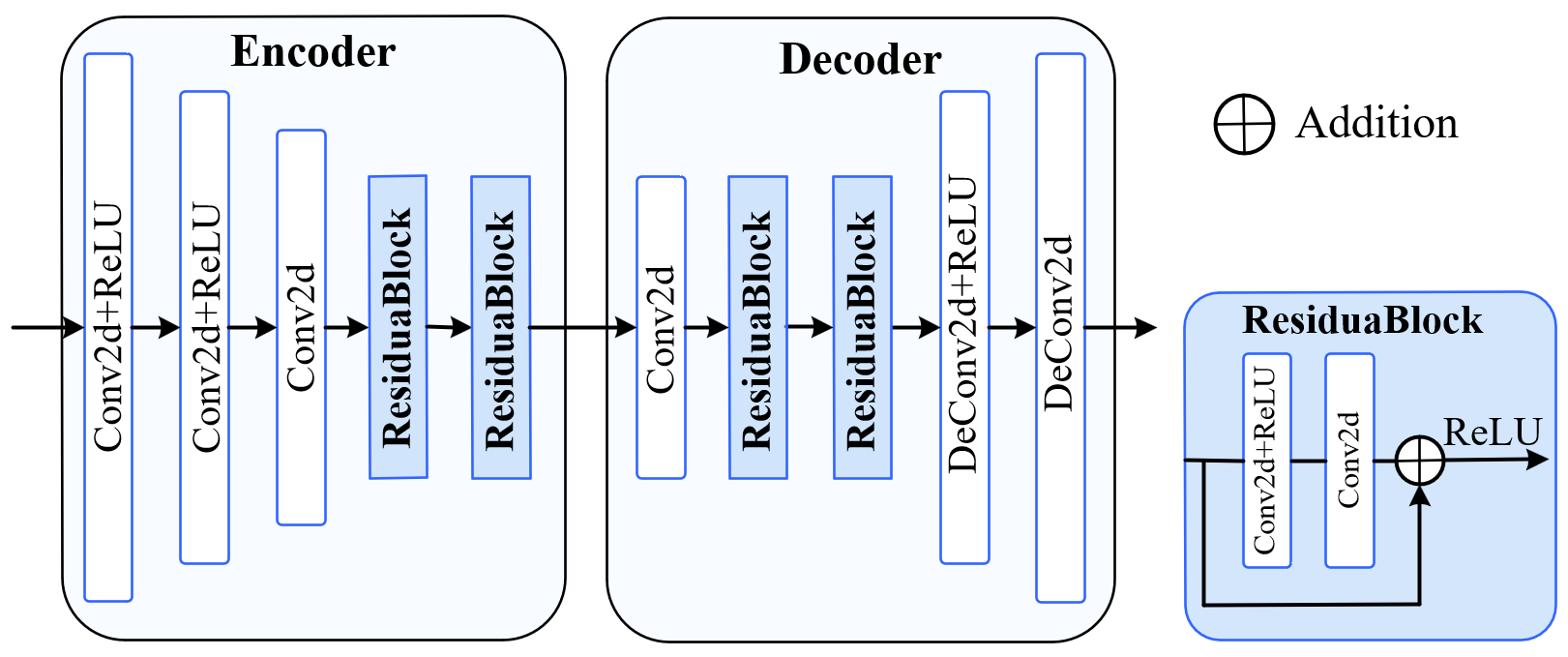}
  \caption{Warp block.}
\end{figure}

\subsubsection{Offset Compensation Branch.} With $j_\theta\left(X_T^i\right)$ as priori, the remaining task is to migrate from $j_\theta\left(X_T^i\right)$ to approach source domain. For this sake, we introduce a module $h_\varphi$ referred to as warp block, to learn target-specific offset compensation for further minimizing the discrepancy from source replay to actual source in terms of reconstruction error. As shown in Figure 3, the warp block is a variant of VQ-VAE \cite{van2017neural}.

Additionally, we apply two scaling factors, $v_S$ and $v_T$, to the source replay branch and the offset compensation branch, respectively, to balance the composition of both branches at the final end of reconstruction, which combines the two streams into a synthetic sequence. During group adaptation, $v_T$ is trainable and $v_S$ is fixed at 1. The final reconstructive output incorporating the two branches can be formulated as
{\begin{equation}
    {\hat{X}}_T^i=v_T\cdot h_\varphi\left(j_\theta\left(X_T^i\right)\right)+v_S\cdot j_\theta\left(X_T^i\right)
\end{equation}}
where $h_\varphi$ denotes the warp block with parameter $\varphi$. The MSE loss to be optimized is:
{\begin{equation}
    {\mathcal L}_{MSE}^T=\frac{1}{N_T} {\sum_{i=1}^{N_T}} \left \|X_T^i-\hat{X}_T^i\right \|^2
\end{equation}}

Aside from the MSE loss in reconstructing target data, we incorporate the improved Tsallis entropy loss \cite{xia2022privacy} into the overall loss when training the warp block and the scaling factor $v_T$. This loss measures the confidence of classifying the target time series without supervision, thereby reducing the uncertainty of the classification.  The overall optimization objective becomes:
{\begin{equation}
\min_{\varphi,v_T}{\mathcal L}_{overall}^T={\mathcal L}_{MSE}^T+\lambda {\mathcal L}_{UR}^T
\end{equation}}where $\lambda$ is a hyperparameter and ${\mathcal L}_{UR}^T$ means improved Tsallis entropy loss, 

\subsection{Instance-wise Adaptation}
Originally, $v_S$ in Equation 2 is fixed at 1. During inference, we impose small perturbation on $v_S$ to test its impact on the stability of classification, and aggregate the classifier’s outputs as weighted sum dependent on the sliding test of stability. Let $\Delta$  represent a small perturbation on $v_S$, and the step-wise perturbation becomes
\begin{equation}
    v_S^j=1\pm j \cdot \Delta, j=0,1,2,\cdots n
\end{equation}
where $n$ refers to the maximum span for parameter variation.

After perturbing the scaling factor, the cosine similarity between classifier’s outputs is computed to measure the confidence of classification, the weighting effect of which is highlighted via softmax in the ensemble of classifier’s outputs.
Let $p_j^i$ represent the output of classifying ${X_{{T}'}^i}$ conditional on the scaling factor ${v_S^j}$ and the cosine similarity $s_j^i=CosSim(p_{j-1}^i,p_{j}^i)$ an indicator to evaluate the stability of the classification when changing $v_S^{j-1}$  to $v_S^j$. After scanning the whole span of $[1-n\cdot\Delta,1+n\cdot\Delta]$, we apply softmax function to obtain the corresponding weights $w_j^i$ as:
\begin{equation}
    w_j^i=\frac{e^{s_j^i}}{\sum_{k=-n}^{n}e^{s_k^i}}
\end{equation}
The classification is finalized as
\begin{equation}
p^i=\sum_{j=-n+1}^{n}{w_j^ip_j^i}
\end{equation}

\section{Experiment}

\subsection{Datasets}
We conduct evaluations on our method using three time series datasets proposed in ADATIME \cite{ragab2023adatime}, all of which are collected from real-world application scenarios, including human activity recognition, sleep stage classification, and machine fault diagnosis.

The three datasets are heterogeneous in terms of time series length and number of channels, and each dataset has a couple of domains with varied feature shift to adapt. The profiles of the three datasets are summarized in Table 2.

\begin{table}[h]
\centering
\caption{Details of three adopted datasets.}
\label{table2}
\begin{tabular}{l|cccc}
\toprule
Dataset & Channel & Length & Class & Domain \\
\midrule
MFD & 1 & 5120 & 3 & 4 \\
SSC & 1 & 3000 & 5 & 20 \\
UCIHAR & 9 & 128 & 6 & 30 \\
\bottomrule
\end{tabular}
\end{table}

\begin{table}[h]
\centering
\caption{Setting of learning rate and epoch.}
\label{table3}
\begin{tabular}{l|ccc}
\toprule
Learning rate, Epoch & Stage 1 & Stage 2 & Stage 3 \\ 
\midrule
MFD                  & 5e-3, 8  & 2e-3, 20 & 5e-3, 8  \\ 
SSC                  & 5e-3, 8  & 2e-3, 15 & 5e-3, 15 \\ 
UCIHAR               & 5e-4, 15 & 5e-3, 15 & 5e-3, 8  \\ 
\bottomrule
\end{tabular}
\end{table}

\subsection{Experimental Setup}
We follow the setting in MAPU \cite{ragab2023source} and use the same classification backbone, allowing our method to be made comparable to MAPU and its baselines. In the literature, MAPU [23] did such verification using three datasets namely MFD, SSC, and UCIHAR from diverse scenarios including mechanical systems, EEG, and wearable sensors, and the SSC data are extremely noisy EEG signals.During pre-training, we reshape and pad the source time series into an image-like form and use it to train the U-net for time series reconstruction. Here thhe MFD, SSC, and UCIHAR data are reshaped to 1×64×80, 1×48×64, and 9×64×64, respectively, with zero padding. The size of the reshaping does not affect the model's performance, as U-Net is powerful in reconstruction. Then, the source time series reconstructed by the frozen U-net is used as the input to train the classification backbone. Finally, we train the warp block and scaling factor $v_T$ using unlabeled target data, during which the pre-trained reconstructor and classification backbone are frozen and access to the source data is prohibited. 
\begin{table*}[b]
  \caption{MF1 score(\%) on MFD benchmark.}
  \label{tab:mfd-benchmark}
  \centering
  \small
  \begin{tabular}{lccccccc}
    \toprule
    Algorithm   & SF         & 0→1            & 1→0            & 1→2            & 2→3            & 3→1            & AVG            \\
    \midrule
    DDC         & \textbf{×} & 74.50          & 48.91          & 89.34          & 96.34          & \textbf{100.0} & 81.82          \\
    DCoral      & \textbf{×} & 79.03          & 40.83          & 82.71          & 98.01          & 97.73          & 79.66          \\
    HoMM        & \textbf{×} & 80.80          & 42.31          & 84.28          & 98.61          & 96.28          & 80.46          \\
    MMDA        & \textbf{×} & 82.44          & 49.35          & \textbf{94.07} & \textbf{100.0} & \textbf{100.0} & 85.17          \\
    DANN        & \textbf{×} & 83.44          & 51.52          & 84.19          & 99.95          & \textbf{100.0} & 83.82          \\
    CDAN        & \textbf{×} & 84.97          & 52.39          & 85.96          & 99.70          & \textbf{100.0} & 84.60          \\
    CoDATS      & \textbf{×} & 67.42          & 49.92          & 89.05          & 99.21          & 99.92          & 81.10          \\
    AdvSKM      & \textbf{×} & 76.64          & 43.81          & 83.10          & 98.85          & \textbf{100.0} & 80.48          \\
    SHOT        & \checkmark & 41.99          & 57.00          & 80.70          & 99.48          & 99.95          & 75.82          \\
    NRC         & \checkmark & 73.99          & 74.88          & 69.23          & 78.04          & 71.48          & 73.52          \\
    AaD         & \checkmark & 71.72          & 74.33          & 78.31          & 90.07          & 87.45          & 80.58          \\
    MAPU        & \checkmark & \textbf{99.43} & 77.42          & 85.78          & 99.67          & 99.97          & 92.45          \\
    \textbf{CT(ours)}     & \checkmark & 99.42          & \textbf{90.96} & 91.05          & 99.34          & \textbf{100.0} & \textbf{96.15} \\
    \bottomrule
  \end{tabular}
\end{table*}

For all the datasets, we use Adam optimizer, and set the batch size to 32, and $\lambda$ to 0.1. Table 3 summarizes the initial learning rate and the number of training epochs for each dataset in each of the three training stages, where the three stages refer to training reconstructor, backbone network, and warp block in order.

In instance adaptation stage, $\Delta$ and n are hyperparameters set as follows: $\Delta$ = 0.1\%, n = 10 for the MFD dataset; $\Delta$ = 0.1\%, n = 8 for the SSC dataset; $\Delta$ = 0.1\%, n = 3 for the UCIHAR dataset. 

We built our model based on Python 3.9 and NVIDIA GeForce RTX 2080Ti GPU. The total number of the parameters of our model is less than 34.74 million, with 34.53 million from reconstructor, 8,170 from the warp block, and 0.2 million from the classification backbone network. The model is computationally efficient to enable 10,293.44 million floating-point operations per second (FLOPS). The relatively small number of parameters and the efficiency make it suitable for deployment in practical applications.

\begin{table*}[h]
\centering
\caption{MF1 score(\%) on SSC benchmark.}
\label{tab:ssc-benchmark}
\small
\begin{tabular}{lccccccc}
\toprule
Algorithm   & SF         & 16→1           & 9→14           & 12→5           & 7→18           & 0→11           & AVG            \\
\midrule
DDC         & \textbf{×} & 55.47          & 63.57          & 55.43          & 67.46          & 54.17          & 59.22          \\
DCoral      & \textbf{×} & 55.50          & 63.50          & 55.35          & 67.49          & 53.76          & 59.12          \\
HoMM        & \textbf{×} & 55.51          & 63.49          & 55.46          & 67.50          & 53.37          & 59.06          \\
MMDA        & \textbf{×} & 62.92          & 71.04          & \textbf{65.11}          & 70.95          & 43.23          & 62.79          \\
DANN        & \textbf{×} & 58.68          & 64.29          & 64.65 & 69.54          & 44.13          & 60.26          \\
CDAN        & \textbf{×} & 59.65          & 64.18          & 64.43          & 67.61          & 39.38          & 59.04          \\
CoDATS      & \textbf{×} & 63.84          & 63.51          & 52.54          & 66.06          & 46.28          & 58.44          \\
AdvSKM      & \textbf{×} & 57.83          & 64.76          & 55.73          & 67.58          & \textbf{55.19} & 60.21          \\
SHOT        & \checkmark & 59.07          & 69.93          & 62.11          & 69.74          & 50.78          & 62.33          \\
NRC         & \checkmark & 52.09          & 58.52          & 59.87          & 66.18          & 47.55          & 56.84          \\
AaD         & \checkmark & 57.04          & 65.27          & 61.84          & 67.35          & 44.04          & 59.11          \\
MAPU        & \checkmark & \textbf{63.85} & \textbf{74.73} & 64.08          & \textbf{74.21} & 43.36          & 64.05          \\
\textbf{CT(ours)}     & \checkmark & 62.99           & 72.15          & 63.38          & 72.33          & 53.3          & \textbf{64.83} \\
\bottomrule
\end{tabular}
\end{table*}

\begin{table*}[h]
\centering
\caption{MF1 score(\%) on UCIHAR benchmark.}
\label{tab:har-benchmark}
\small
\begin{tabular}{lccccccc}
\toprule
Algorithm   & SF         & 2→11           & 12→16          & 9→18           & 6→23           & 7→13           & AVG            \\
\midrule
DDC         & \textbf{×} & 60.00          & 66.77          & 61.41          & 88.55          & 77.29          & 75.67          \\
DCoral      & \textbf{×} & 67.20          & 64.58          & 54.38          & 89.66          & 90.46          & 84.10          \\
HoMM        & \textbf{×} & 83.54          & 63.45          & 71.25          & 94.97          & 91.41          & 84.10          \\
MMDA        & \textbf{×} & 72.91          & 74.64          & 62.62          & 91.14          & 90.61          & 81.40          \\
DANN        & \textbf{×} & 98.09          & 62.08          & 70.70          & 85.60          & 93.33          & 84.97          \\
CDAN        & \textbf{×} & 98.09          & 61.20          & 71.30          & 96.73          & 93.33          & 86.79          \\
CoDATS      & \textbf{×} & 86.65          & 61.03          & 80.51          & 92.08          & 92.61          & 85.47          \\
AdvSKM      & \textbf{×} & 65.74          & 60.52          & 53.25          & 79.63          & 88.89          & 74.67          \\
SHOT        & \checkmark & \textbf{100.0} & 70.76          & 70.19          & \textbf{98.91}          & 93.01          & 86.57          \\
NRC         & \checkmark & 97.02          & 72.18          & 63.10          & 96.41          & 89.13          & 83.57          \\
AaD         & \checkmark & 98.51          & 66.15          & 68.33          & 98.07 & 89.41          & 84.09          \\
MAPU        & \checkmark & \textbf{100.0} & 67.96          & 82.77          & 97.82          & \textbf{99.29} & 89.57          \\
\textbf{CT(ours)}     & \checkmark & \textbf{100.0} & \textbf{86.88}          & \textbf{87.45} & 95.47          & 91.04          & \textbf{92.17} \\
\bottomrule
\end{tabular}
\end{table*}

\subsection{Performance Evaluation}
Adopting the baselines from MAPU, we compare our method with the conventional UDA methods, including DDC \cite{tzeng2014deep}, DCoral \cite{sun2017correlation}, HoMM \cite{chen2020homm}, MMDA \cite{rahman2020minimum}, DANN \cite{ganin2016domain}, CDAN \cite{long2018conditional}, CoDATS \cite{wilson2020multi}, and AdvSKM \cite{liu2021adversarial}, as well as the four SFDA methods, including SHOT \cite{liang2020we}, NRC \cite{yang2021exploiting}, AaD \cite{yang2022attracting}, and MAPU \cite{ragab2023source}.

We compute the MF1 score for each of the five domain transferring scenarios, and the average over all scenarios. The results are shown in Tables 4, 5, and 6 for the three datasets, respectively, where SFDA against UDA is marked as "$\surd$" or "×", respectively, and CT refers to our method, say, compositional transferring.

Compared with the most competitive ones, our method outperforms all baselines on the 3 datasets, leading to the SOTA performance of 96.15\%, 64.83\%, and 92.17\% in terms of MF1 score, which means 3.7\%, 0.78\%, and 2.6\% improvement on MFD, SSC, and UCIHAR benchmark. It is worth noting that our method has made significant progress in some scenarios: For example, for the 1→0 scenario of MFD and the 12→16 scenario of UCIHAR, our method improves the MF1 score by 13\% and 12\%, respectively. At the same time, our method performs pervasively well in the remaining tests, close to the best performing baselines. These account for why our method outperforms the baselines in the overall sense.

% \section{ABLATION STUDY}
\section{Ablation Study}

% {\bfseries Rationality of Model Structure. }
\subsection{Model Structure}
There are two reconstruction modules in our method: The pre-trained source domain reconstruction module and the warp block for group adaptation. As shown in Table 7, we test the impact of the implementation of the two modules, where the former one refers to the source domain reconstructor, and the latter one is the warp block. The number of parameters of U-net and AE are 34.53 million and 8170, respectively.
\begin{table}[ht]
  \centering
  \begin{minipage}[t]{0.5\textwidth} % 设置第一张表格的宽度为 48% 的文本宽度
    \centering
    \caption{Configurations of two reconstruction modules.}
    \label{table7}
    \small % 调整表格内容的字体大小为 9 号字体
    \begin{tabular}{l|ccc}
      \toprule
      MF1 score(\%) & {MFD} & {SSC} & {UCIHAR} \\
      \midrule
      AE + AE & 89.56 & 57.63 & 26.65 \\
      AE + U-net & 91.03 & 56.75 & 29.19 \\
      U-net + U-net & 91.9 & 60.94 & 87.12 \\
      U-net + AE     & \textbf{96.15} & \textbf{64.83} & \textbf{92.17} \\
      \bottomrule
    \end{tabular}
  \end{minipage}
  \hfill % 添加一些水平间距
  \begin{minipage}[t]{0.48\textwidth} % 设置第二张表格的宽度为 48% 的文本宽度
    \centering
    \caption{Advantage of the two-branch design.}
    \label{table8}
    \small % 调整表格内容的字体大小为 9 号字体
    \begin{tabular}{l|ccc}
      \toprule
      MF1 score(\%) & {MFD} & {SSC} & {UCIHAR} \\
      \midrule
      CT w/o SR-branch & 73.26 & 50.97 & 48.25 \\
      CT w/o OC-branch & 77.84 & 59.1 & 86 \\
      Full CT     & \textbf{96.15} & \textbf{64.83} & \textbf{92.17} \\
      \bottomrule
    \end{tabular}
  \end{minipage}
\end{table}

% \begin{table}[h!]
% \centering
% \caption{Configurations of two reconstruction modules.}
% \label{table6}
% \begin{tabular}{l|ccc}
% \toprule
% MF1 score(\%) & {MFD} & {SSC} & {UCIHAR} \\
% \midrule
% AE + AE & 89.56 & 57.63 & 26.65 \\
% AE + U-net & 91.03 & 56.75 & 29.19 \\
% U-net + U-net & 91.9 & 60.94 & 87.12 \\
% U-net + AE     & \textbf{96.15} & \textbf{64.83} & \textbf{92.17} \\
% \bottomrule
% \end{tabular}
% \end{table}

From the table, we can see that the best performance is achieved by using U-net followed by AE. This is because a larger number of parameters are required to fit the source domain distribution, which forms the basis for further transferring, while warp block is only responsible for compensating the offset on the basis of source model replay. Here, the different fitting power of U-net against that of AE arises from the different numbers of parameters.

% \begin{table}[h!]
% \centering
% \caption{Advantage of the two-branch design.}
% \label{table6}
% \begin{tabular}{l|ccc}
% \toprule
% MF1 score(\%) & {MFD} & {SSC} & {UCIHAR} \\
% \midrule
% CT w/o SR-branch & 73.26 & 50.97 & 48.25 \\
% CT w/o OC-branch & 77.84 & 59.1 & 86 \\
% Full CT     & \textbf{96.15} & \textbf{64.83} & \textbf{92.17} \\
% \bottomrule
% \end{tabular}
% \end{table}

To validate the two-branch design, we provide comparative experiments in Table 8, where SR represents source replay, OC denotes offset compensation, and CT means compositional transferring. It can be seen that the performance degradation without SR branch or OC branch is significant. Note that CT without OC-branch is actually the sole source replay, whose performance provides a solid base for full CT as shown in the tabel. This explains why source replay based on pre-trained model is essential.

\subsection{Loss Selection.}
When migrating the target domain at group level, we use an uncertainty reduction loss ${\mathcal L}_{UR}^T$ in addition to MSE loss ${\mathcal L}_{MSE}^T$. We test its contribution in Table 9. It augments our model.

\subsection{Instance-wise Adaptation (IA)}

We compare cosine similarity to entropy based measurement in testing stability of classification. As shown in Table 10, our metric outperforms entropy. The sensitivity of the hyperparameter setting of our method is shown in Figure 4. It can be seen that our method is ubiquitously stable.

% \begin{table}[h!]
% \centering
% \caption{Performance improvement with ${\mathcal L}_{UR}^T$.}
% \label{table6}
% \begin{tabular}{l|ccc}
% \toprule
% MF1 score(\%) & {MFD} & {SSC} & {UCIHAR} \\
% \midrule
% \small${\mathcal L}_{MSE}^T$ & 94.41 & 63.31 & 90.25 \\
% \small${\mathcal L}_{MSE}^T+{\mathcal L}_{UR}^T$     & \textbf{96.15} & \textbf{64.83} & \textbf{92.17} \\
% \bottomrule
% \end{tabular}
% \end{table}
\begin{table}[h!]
  \centering
  \begin{minipage}[t]{0.48\textwidth} % 设置第一张表格的宽度为 48% 的文本宽度
    \centering
    \caption{Performance improvement with ${\mathcal L}_{UR}^T$.}
    \label{table9}
    \small % 调整表格内容的字体大小为 9 号字体
    \begin{tabular}{l|ccc}
      \toprule
      MF1 score(\%) & {MFD} & {SSC} & {UCIHAR} \\
      \midrule
      ${\mathcal L}_{MSE}^T$ & 94.41 & 63.31 & 90.25 \\
      ${\mathcal L}_{MSE}^T+{\mathcal L}_{UR}^T$     & \textbf{96.15} & \textbf{64.83} & \textbf{92.17} \\
      \bottomrule
    \end{tabular}
  \end{minipage}
  \hfill % 添加一些水平间距
  \begin{minipage}[t]{0.48\textwidth} % 设置第二张表格的宽度为 48% 的文本宽度
    \centering
    \caption{Advantage of cosine similarity.}
    \label{table10}
    \small % 调整表格内容的字体大小为 9 号字体
    \begin{tabular}{l|ccc}
      \toprule
      MF1 score(\%) & {MFD} & {SSC} & {UCIHAR} \\
      \midrule
      CT w/o IA & 95.13 & 64.28 & 91.93 \\
      \midrule
      IA by Entropy & 96.1 & \textbf{64.83} & 91.93 \\
      IA by CosSim     & \textbf{96.15} & \textbf{64.83} & \textbf{92.17} \\
      \bottomrule
    \end{tabular}
  \end{minipage}
\end{table}

% {\bfseries Instance-wise adaptation (IA). }

% \begin{table}[h!]
% \centering
% \caption{Advantage of cosine similarity.}
% \label{table6}
% \begin{tabular}{l|ccc}
% \toprule
% MF1 score(\%) &{MFD} & {SSC} & {UCIHAR} \\
% \midrule
% CT w/o IA & 95.13 & 64.28 & 91.93 \\
%  \hline
% IA by Entropy & 96.1 & \textbf{64.83} & 91.93 \\
% IA by CosSim     & \textbf{96.15} & \textbf{64.83} & \textbf{92.17} \\
% \bottomrule
% \end{tabular}
% \end{table}

% \begin{figure}[h]
%   \centering
%   \includegraphics[width=0.4\linewidth]{pic4}
%   \caption{Sensitivity of IA to $\Delta$ and n.}
% \end{figure}
\begin{figure}[h]
  \centering
  \begin{minipage}[t]{0.45\textwidth} % 设置第一张图片的宽度为 30% 的文本宽度
    \centering
    \includegraphics[width=\linewidth]{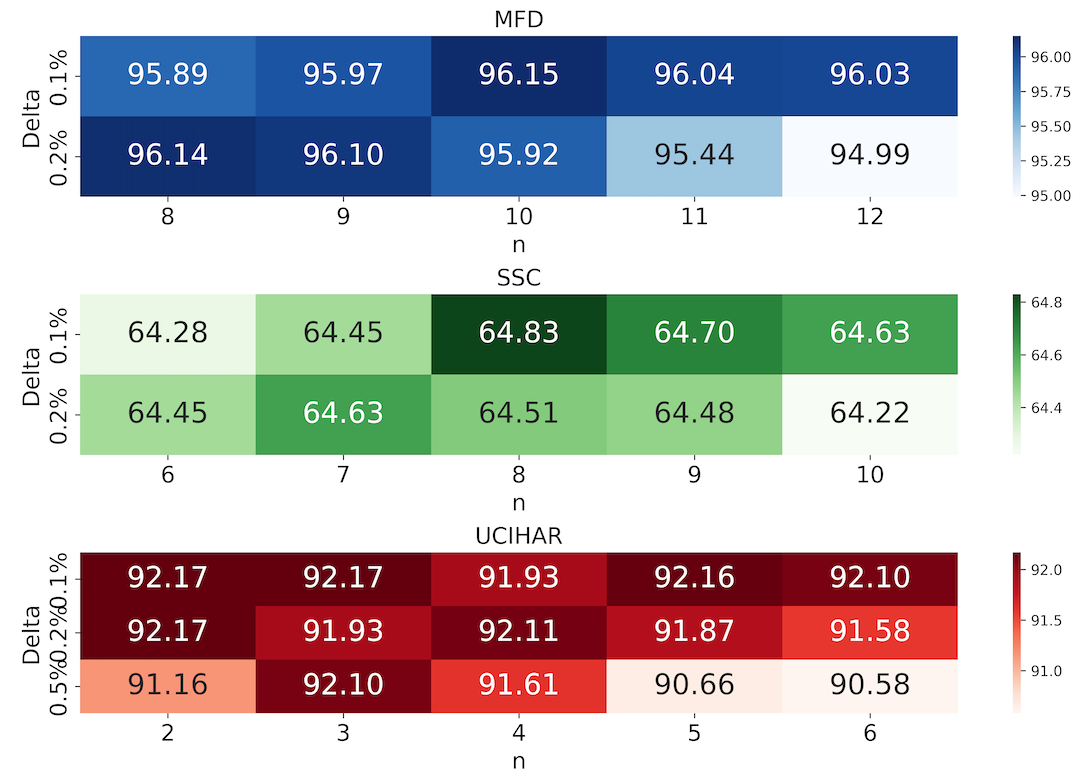}
    \caption{Sensitivity of IA to $\Delta$ and n.}
    \label{fig:pic4}
  \end{minipage}
  \hfill % 添加一些水平间距
  \begin{minipage}[t]{0.45\textwidth} % 设置第二张图片的宽度为 30% 的文本宽度
    \centering
    \includegraphics[width=\linewidth]{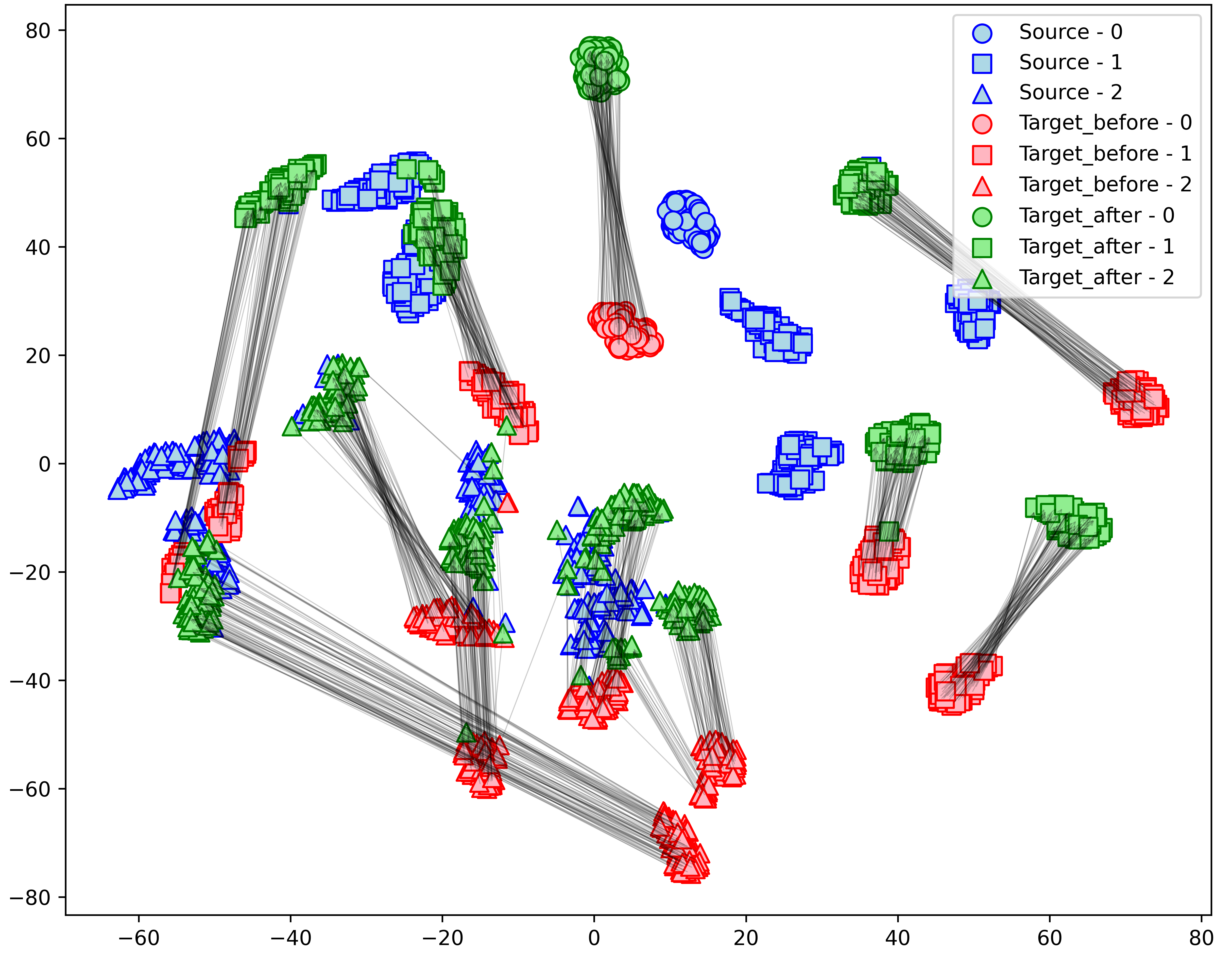}
    \caption{Visualization of the three classes of MFD data for 0→1 domain transfer.}
    \label{fig:pic5}
  \end{minipage}
\end{figure}

% \section{INTERPRETIVE EXPERIMENTS}
\section{Interpretive Experiments}

% {\bfseries Feature visualization. }
\subsection{Feature Visualization}
As shown in Figure 5, we compare the target data before and after adaptation using our method, in contrast to the source data in the latent space as visualized by t-SNE with dimensionality reduction. Note that the latent features obtained by using the classification backbone $g$ in Figure 2 are illustrated in different shapes in Figure 5 to denote their belongings to 3 classes, where circle, square, and triangle represent the data instances of class 0, class 1, and class 2 in the MFD data, respectively. Further, we assign blue, red, or green color to each data instance to denote the source data, the target data before adaptation, and the target data after adaptation, where the black lines indicate the course of migrating the target data from their original positions to the destinations in the latent space to undergo domain adaptation. It is apparent that the target data becomes closer to the source data after adaptation.

% \begin{figure}[h]
%   \centering
%   \includegraphics[width=\linewidth]{pic5}
%   \caption{Visualization of the three classes of MFD data for 0→1 domain transfer.}
% \end{figure}

\begin{figure}[h]
  \centering
  \includegraphics[width=0.5\linewidth]{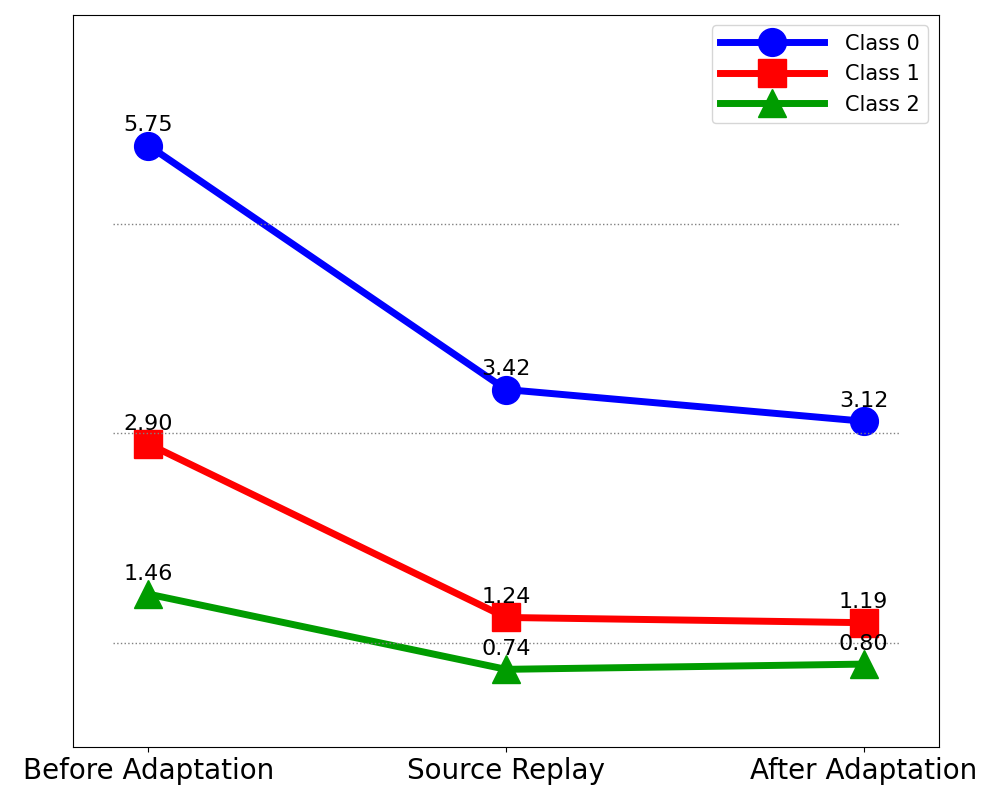}
  \caption{The average distance from the MFD target data to their nearest neighbors in the source domain when transferring from domain 0 to domain 1. The three stages in the curve marked with solid dots correspond with no adaptation, source replay based adaptation, and full adaptation.}
\end{figure}

% {\bfseries Statistical interpretation. }
\subsection{Statistical Interpretation}
From a statistical point of view, we average the distances from the target features to their nearest neighbors in source domain before and after adaptation, and the middle stage of source replay, respectively, where the features of time series are obtained by using the classification backbone $g$ as illustrated in Figure 2. In Figure 6, source replay is a critical step to approach the full adaptation, which shortens the domain gap significantly. The full adaptation with additional offset compensation can further remove the gap out.

\section{Conclusion}

This study is focused on SFDA for time series classification. Based on observing the commitment of compositional transferability to domain adaptation, we propose the compositional design combining source replay and offset compensation, with online instance-wise rescaling. SOTA performance has been achieved experimentally.

\bibliographystyle{plainnat} % 指定参考文献样式
\bibliography{CT}

\end{document}